\documentclass[journal]{IEEEtran}
\ifCLASSINFOpdf
\else
\fi
\usepackage{cite}
\usepackage{amsmath,amssymb,amsfonts}
\usepackage{bm}
\usepackage{algorithm}
\usepackage{algpseudocode}
\usepackage{dblfloatfix}
\algdef{SE}
[CLASS]
{Class}
{EndClass}
[1]
{\textbf{class} \textsc{#1}}
{\textbf{end class}}

\usepackage{graphicx}
\usepackage{textcomp}
\usepackage{xcolor}
\usepackage{multirow}
\usepackage[math]{cellspace}
\usepackage{comment}

\begin{document}
%
\title{Adapting to Covariate Shift in Real-time \\ by Encoding Trees with Motion Equations}

%
%
%

\author{Tham~Yik~Foong,
        Heng~Zhang,
        Mao~Po~Yuan,
        and~Danilo~Vasconcellos~Vargas
\thanks{Danilo Vasconcellos Vargas are with the Department of Information Science and Technology at Kyushu University, Japan, and The University of Tokyo, Tokyo, Japan.}%
\thanks{Tham Yik Foong, Heng Zhang, and Mao Po Yuan are with the Department of Information Science and Technology at Kyushu University, Japan.}%
\thanks{This work was supported by JST, ACT-I Grant Number JP-50243 and JSPS KAKENHI Grant Number JP20241216. Tham Yik Foong and Heng Zhang are supported by JST SPRING, Grant Number JPMJSP2136.}%
}

%
%

\markboth{Journal of \LaTeX\ Class Files,~Vol.~14, No.~8, August~2015}%
{Shell \MakeLowercase{\textit{et al.}}: Bare Demo of IEEEtran.cls for IEEE Journals}
%



\maketitle

\begin{abstract}
Input distribution shift presents a significant problem in many real-world systems.
Here we present Xenovert, an adaptive algorithm that can dynamically adapt to changes in input distribution.
It is a perfect binary tree that adaptively divides a continuous input space into several intervals of uniform density while receiving a continuous stream of input.
This process indirectly maps the source distribution to the shifted target distribution, preserving the data's relationship with the downstream decoder/operation, even after the shift occurs.
In this paper, we demonstrated how a neural network integrated with Xenovert achieved better results in 4 out of 5 shifted datasets, saving the hurdle of retraining a machine learning model.
We anticipate that Xenovert can be applied to many more applications that require adaptation to unforeseen input distribution shifts, even when the distribution shift is drastic.
\end{abstract}


%
\IEEEpeerreviewmaketitle

\section{Introduction}

Changes are inevitable in nature; a system that is built around the assumption that the environment is static often fails when the input distribution shifts.
For instance, a phenomenon called covariate shift happens when the input data used to train a model differs from the input data encountered in testing \cite{bib1D11, bib1D12, bib1D7, bib1D5}.
Research examining covariate shift has shown linear trends between performance on
shifted and unshifted test distributions, which degrade the performance of the application \cite{bib1D26, bib1D27}.
The earliest solutions proposed to combat covariate shift involved adjusting the importance of training instances to reduce the discrepancy between training and testing distributions \cite{bib1D28, bib1D14, bib1D29, bib1D32}. 
Subsequent methods involved finding the differences in densities between two functions and then utilizing importance-weighted learning methods to determine the parameters for training the models \cite{bib1D30, bib1D31}. 
Additionally, more effective shift detection methods have also been developed to enable counteractions \cite{bib1D10}.

Currently, machine learning approaches still fall short in terms of adaptability when handling the problem of covariate shift in an online fashion \cite{bib1D7}. 
Most of the methods proposed are limited by the assumption that the training and testing distributions should overlap to some extent \cite{bib1D8}.
Often, the common workaround during the event of a covariate shift is to retrain the model on the shifted dataset \cite{bib1D9, bib1D10}.
This leads to a significant increase in the cost of maintaining a machine learning model (i.e., the cost is directly proportional to the scale of the dataset and the frequency of the shifting events).
Despite being a common phenomenon that is ubiquitous in many scenarios, these issues are not properly addressed by an adaptive algorithm.

\begin{figure*}[htbp]
\centerline{\includegraphics[width = 18cm]{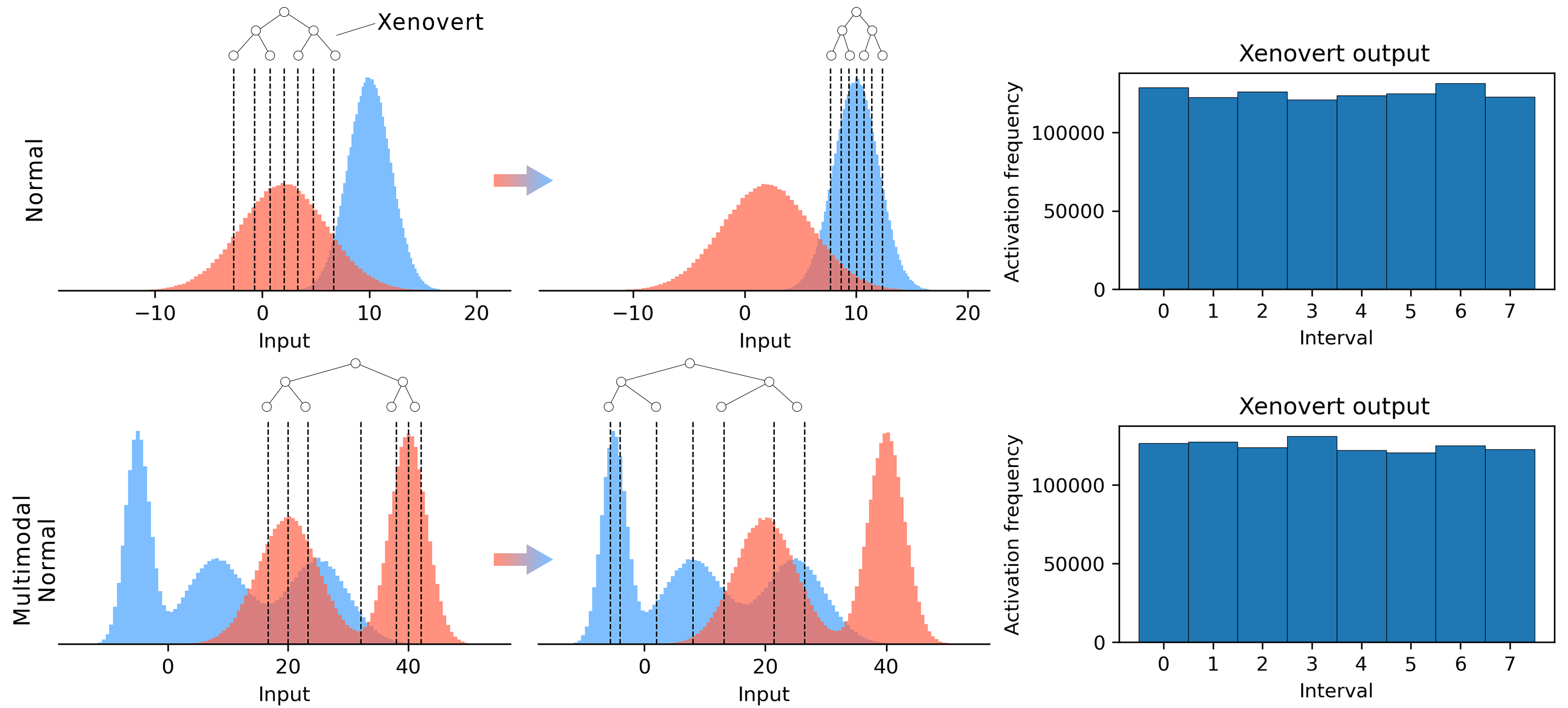}}
\caption{Xenovert's adaptation from source distribution (red) to target distribution (blue). The histogram displays the frequency of interval activation, indicating that all intervals have nearly identical input frequencies. Thus, this shows that the quasi-quantiles can divide the input distribution uniformly.}
\label{fig1}
\end{figure*}

To mitigate these issues, we introduce a parametric algorithm named Xenovert. 
It uses a perfect binary tree to dynamically partition a continuous input space into multiple intervals, each quantized to represent different degrees of value.
It achieves adaptation by adjusting the sizes of these intervals while processing input sequentially to avoid imbalances among them.
This process maps the source distribution to a shifted target distribution via an intermediate representation, which is a set of intervals (see Fig.~\ref{fig1}).
Our experimental analysis of distribution shifts has demonstrated the novelty of our method for continuously adapting to drastic distribution shifts.
Furthermore, we integrate the Xenovert with neural network to address covariate shift, demonstrating its practical application on shifted datasets.
Without any retraining required, inputs processed by the Xenovert can retain the mapping from the original training input to the shifted input, essentially allowing the downstream neural network to adapt to the shifted input distribution.
The algorithm has been validated under three specific shifting conditions: instant, gradual, and recurring, using different types of probability distributions.

\hfill \break
\noindent
To summarize, this paper makes the following contributions:
\begin{itemize}
  \item We have developed an adaptive algorithm capable of managing drastic shifts in the input distribution, even in cases where there is no overlap between the source and target distributions, a topic yet to be extensively explored in the field of machine learning.
  \item We have integrated our algorithm with a neural network to enhance its robustness against covariate shift and achieved better in 4 out of 5 shifted datasets.
  \item Through investigations of Xenovert with different types of probability distributions and shifting types, we provide an extensive analysis of our methods using simulation data. 
\end{itemize}


\section{Related work}

Numerous algorithms have been developed to either detect or adapt to distribution shifts.
Commonly, statistical hypothesis testing methods are employed for detection purposes.
For instance, the Kolmogorov-Smirnov Test identifies when two distributions differ, yet it lacks the ability to provide detailed quantification of the differences.
Conversely, the shift function and its variants, as in \cite{bib1D3}, are adept at detecting and quantifying the degree of shifting between two distributions.
To visualize and understand multivariate distribution shifts, an algorithm called Relative Density Clouds based on k-nearest neighbor density estimates was proposed. This method is particularly useful for identifying and interpreting complex patterns of changes in distribution characteristics, such as location, scale, and variation, which aligns with the challenges of detecting shifts in multivariate distributions \cite{del2023relative}.

Nevertheless, detecting a covariate shift alone is not sufficient.
For models to make accurate predictions, adapting to these shifts is essential.
There are algorithms that use moving-quantiles, akin to those in this work, to adapt to recent changes in data streams.
However, these methods rely typically on predefined parameters, such as sliding windows, sketches (compact representations in data stream processing), or frequency counters \cite{bib1D1, bib1D2, bib1D22, bib1D21}.
A current benchmark for adapting to covariate shifts is the Dynamic Importance Weighting (DIW) algorithm, which has demonstrated effectiveness in handling these shifts in datasets \cite{bib1D8}.
However, DIW, which utilizes weight estimation methods such as Kernel Mean Matching, Least-Squares Importance Fitting, and Direct Importance Estimation, encounters limitations in handling drastic shifts.
This is primarily due to these methods' heavy reliance on the overlap between training and test distributions \cite{bib1D6, bib1D7, bib1D8, bib1D24, bib1D25}.
The accuracy of weight estimations diminishes when distributions have minimal overlap or when relationships between variables undergo substantial changes, leading to reduced effectiveness under severe distribution shift conditions.
More recently, \cite{zhang2023adapting} addresses the challenge of continuous covariate shift by using an online method for density ratio estimation under conditions of data scarcity. Drastic shifts, however, are under-explored, as the method is aimed at minimizing risk while training predictors over time, especially in scenarios with continuously shifting test data distributions. Additionally, the concept of online label shift has been explored in recent studies \cite{bai2022adapting, yan2023online}.

\section{Methodology}

\subsection{Xenovert}

\begin{figure}[htbp]
\centerline{\includegraphics[width = 8cm]{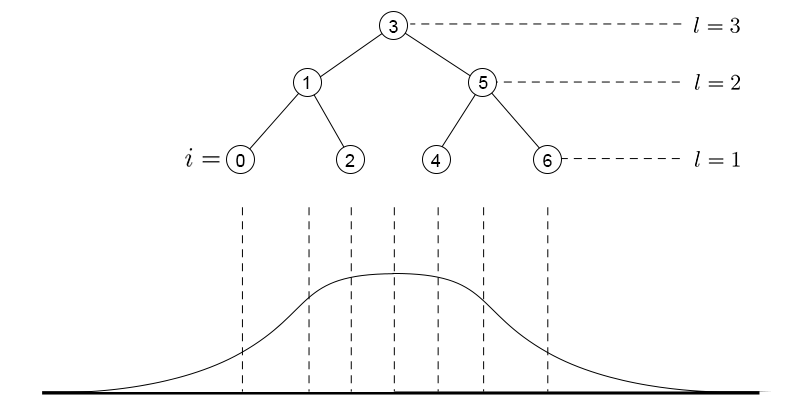}}
\caption{Illustration of indexing and level structure in Xenovert. Xenovert divides the input using a tree structure with each node setting boundaries into quasi-quantile, with leaf nodes indexing these quasi-quantiles. The index of quasi-quantiles, $i = \{0, 1, \cdots, n\}$ are sorted from left to right in ascending order regardless of the hierarchy. The levels are sorted in descending order in a top-down manner.}
\label{fig7}
\end{figure}

\begin{figure*}[htbp]
\centerline{\includegraphics[width = 18cm]{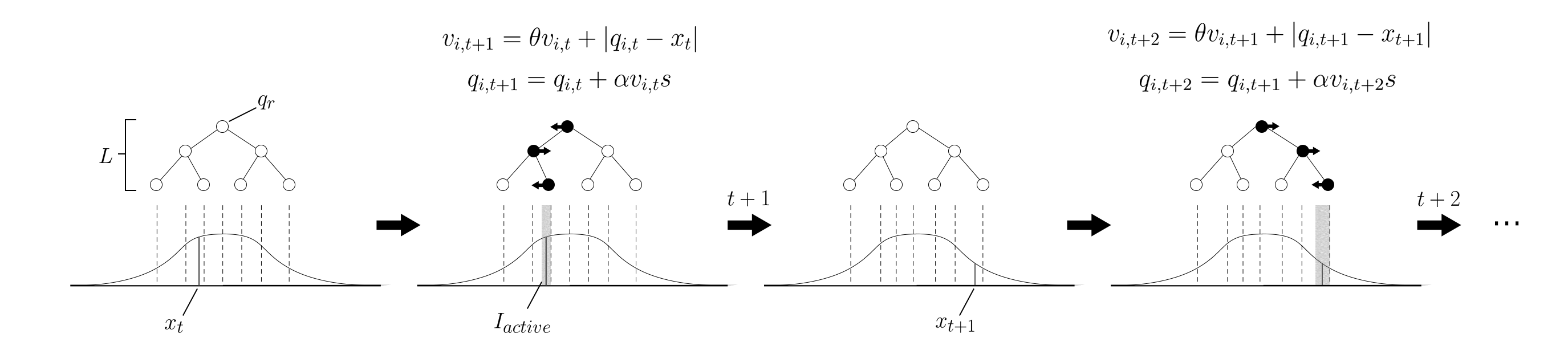}}
\caption{The illustration of Xenovert's update function. Xenovert is a perfect binary tree that uses its nodes to represent quasi-quantiles $q$, which divide the input space into equal-sized intervals.
In detail, when given an input $x$ at time step $t$, the chosen quasi-quantiles approximate towards the input with the update function $q_{i,t+1} = q_{i,t} + \alpha v_{i,t+1}s$; Where $v$ is a velocity term, $alpha$ is the learning rate, $\theta$ is the velocity decay, and $s=-1$ if $q-x>0$, otherwise, $s=1$. Starting from the root quasi-quantiles $q_r$, we select the `left' child node if $x < q_r$ (or $x < q$, when it is not a root node); otherwise, select the `right' child node. The selected child node is marked in black. The gray shade is the interval that the input falls into. If the input distribution remains constant, the quasi-quantiles converge to an equilibrium state, eventually ensuring that all intervals contain an equal concentration of inputs.
}
\label{fig9}
\end{figure*}

We propose the Xenovert, designed to adaptively segment a continuous input space into multiple uniform intervals. 
The workflow of the Xenovert is outlined as follows:
First, we define the Xenovert as a perfect binary tree, denoted as $T$, with a total $L$ level.
Followed by that, each node in the Xenovert is conceptualized as a quasi-quantile, represented by a scalar value $q_i \in \mathbb{R}$, with the series ranging from $q_0$ to $q_N$, where $N = 2^L - 1$. 
The root quasi-quantile, or root node, is represented as $q_r = \frac{N+1}{2}$.
The correspondence between the indexing of these quasi-quantiles and the levels of the binary tree is illustrated in Figure~\ref{fig7}.
As the level $L$ dictates the quasi-quantile count, it is chosen to optimize between detailed input representation or adaptability, depending on the task requirements.

The function for updating these quasi-quantiles is first conceptualized as a Widrow-Hoff function for scalar value (quasi-quantile's value). 
Therefore, we can omit the cross-product with the input as the error already reflects the input's magnitude and direction:

\begin{equation}
\label{eq:0}
q_{i,t+1} = q_{i,t} + \alpha (q_{i,t}-x_t)
\end{equation}

\noindent where $\alpha$ is the learning rate and $x_t$ is the current input, as well as the expected value of $q_{i,t}$.
The learning rule dictates that whenever an input $x$ is introduced, the quasi-quantiles $q_{i,t+1}$ are updated to gradually approach the value of this input. 
However, continuous input can cause fluctuations around the expected value and fail to maintain stability. 
To address this, we redefined the error term using a motion equation, resulting in the final quasi-quantiles update function described by:

\begin{equation}
\label{eq:1}
s = \left\{\begin{matrix}
-1 & , q-x>0 \\ 
1 & , otherwise
\end{matrix}\right.
\end{equation}

\begin{equation}
\label{eq:2}
v_{i,t+1} = \theta v_{i,t} + |q_{i,t}-x_t|
\end{equation}

\begin{equation}
\label{eq:3}
f(q_i) = q_{i,t+1} = q_{i,t} + \alpha v_{i,t+1}s
\end{equation}

\noindent where $v$ is a velocity term and $\theta$ is the velocity decay.
We initialize $\alpha$ to $1e-05$, $v$ to $0$ and $\theta$ to $0.99$.
Over time, this motion equation enables the quasi-quantiles to approximate the average of their respective intervals.
The velocity term $v$ serves as an damping or momentum factor for each update, modulating the update rate of quasi-quantiles to prevent excessive adjustments.
In other words, small updates occur when the selected quasi-quantile is close to the inputs, while larger updates occur if the quasi-quantile is far from the subsequent inputs. The term $s$ in Eq.~\ref{eq:1} determines the update direction.

The update occurs recursively from the root node to one of the leaf nodes.
Only the root node and the selected nodes will get updated, and the selection criteria for the selected child node (quasi-quantiles) on each level are defined by the selection function:

\begin{equation}
\label{eq4}
f(q_j)| \forall l <= L, j = \left\{\begin{matrix} & i-2^{l-2}, x<q_i\\  & i+2^{l-2}, x\geq q_i\end{matrix}\right.  
\end{equation}

\noindent with $l$ being the current level in the binary tree, $i$ the index of the current selected node (quasi-quantile), and $j$ the next selected node.
At each time step, the algorithm updates the selected quasi-quantiles on every level ($L$ number of quasi-quantiles).
Assuming the input distribution does not change, these quasi-quantiles will uniformly distribute across the input space over time.
If a distribution shift occurs, the quasi-quantiles actively adapt to this new distribution until it stabilizes. 
Figure~\ref{fig9}A illustrates the Xenovert's update and selection functions.

To yield an output that is adapted to the distribution shift, Xenovert converts input into quantized output $y$ by:

\begin{equation}
C(q_j), j = \left\{\begin{matrix} & i-2^{l-2}, x<q_i\\  & i+2^{l-2}, x\geq q_i\end{matrix}\right. , | \forall l <= L 
\end{equation}

The convert function $C(q_j)$ is a recursive function that iterates for $L$ times, and then returns the output value $y$, which equals $j$.


\begin{figure*}[htbp]
\centerline{\includegraphics[width = 18cm]{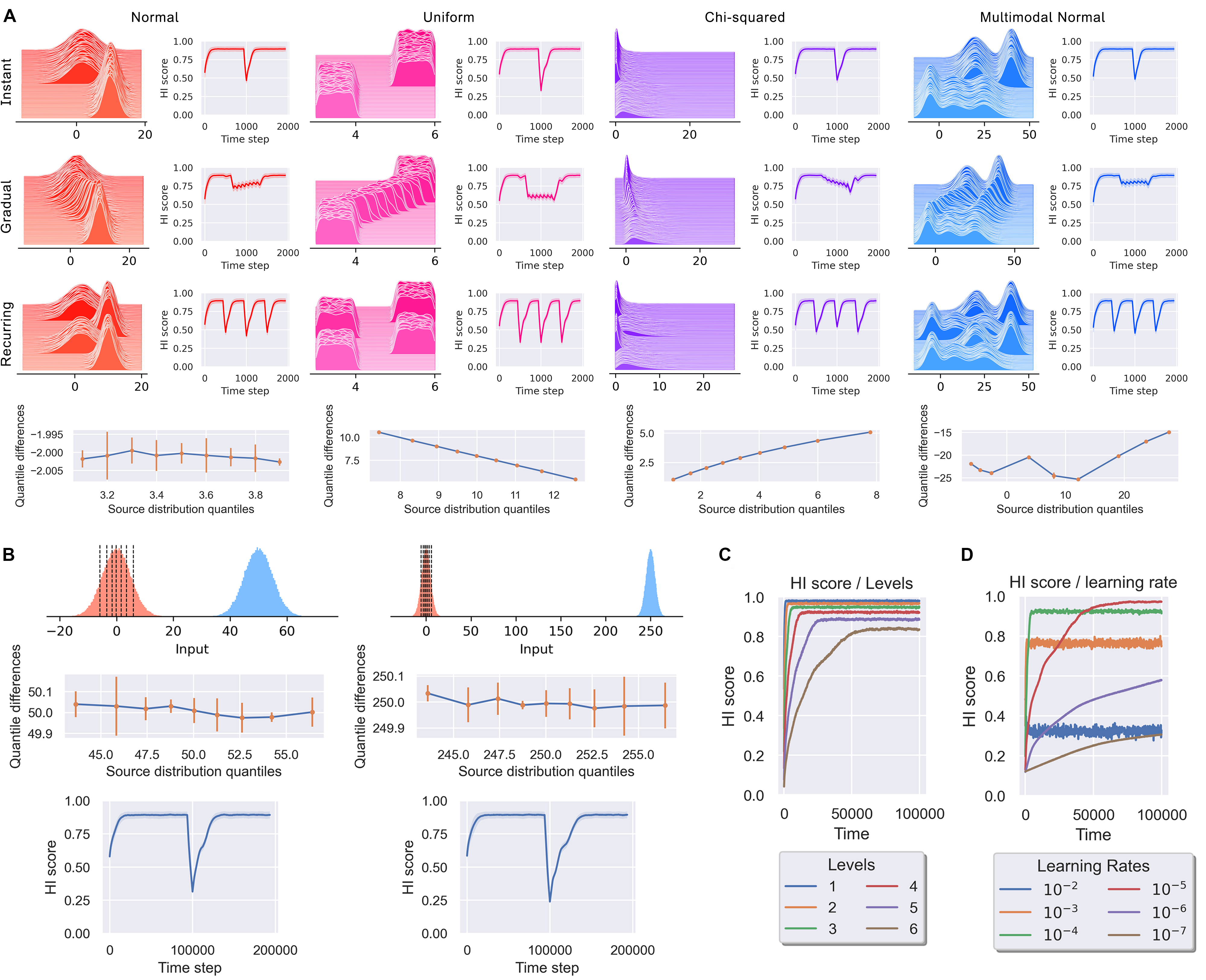}}
\caption{Simulation result of distributions with three types of shifting. (A) The ridgeline plot shows the shifting from source distribution to target distribution (bottom to top). The bottom row shows the degree of shifting in source and target distribution. Based on the HI score learning curve, we observed a general trend where the HI score drops when shifting starts to occur and returns to a near-optimum score when Xenovert adapts to the new distribution. (B) The adaptation of Xenovert is invariant to the degree of shifting. Firstly, we experimented with Xenovert's adaptation using inputs drawn from a normal distribution, $\mathcal{N}(0, 5)$ and $\mathcal{N}(50, 5)$ as the source and target distributions, respectively. Followed by that, we experimented with another set of distributions, $\mathcal{N}(0, 5)$ and $\mathcal{N}(250, 5)$. The HI score learning curve, however, remains almost identical in both experiments, proving its invariance to the degree of shifting. (C) The higher the total level, the lesser the data compression and information lost, but the lower the HI score. (D) The stability-adaptability trade-off, where the higher the learning rate, the lower the stability but the faster the adaptability.}
\label{fig4}
\end{figure*}

\section{Analyzing Xenovert on \\ Univariate Distribution Shift}

\subsection{Experiment setup}

In this section, we evaluated the performance of Xenovert on various types of distributions with three \textit{types of shifts}.
The \textit{type of shift} indicates whether the transition from source distribution to target distribution takes place instantly or gradually over time.
Besides, in the case of recurring shifts, inputs are drawn seasonally or cyclically from the source distribution and target distribution.
Fig.~\ref{fig4} shows the experiment setup that evaluates Xenovert on Uniform, Normal, Multimodal Normal, Chi-square, and Multivariate Normal distribution with instant, gradual, and recurring shifting.
In each experiment, we generate a pair of source distributions and target distributions.
Both source and target distributions are defined with different parameters.
For instance, in Normal distribution tasks, the source distribution is $\mathcal{N}(2, 4)$ while the target distribution is $\mathcal{N}(10, 2)$.
Inputs drawn from the source distribution are treated as input before the distribution shift occurs, while inputs drawn from the target distribution are treated as input after the distribution shift occurs.

The degree of shifting in this experiment is measured using a shift function. 
The shift function quantifies how much two distributions differ; compare this to a significant Kolmogorov–Smirnov test, where it only informs that the two distributions differ, not how much.
To verify whether the inputs drawn from the target distribution are uniformly distributed by the quantiles, we use the histogram intersection (HI) score to compare the prediction of Xenovert to a uniform distribution.
The HI score is a real value ranging from 0 to 1 that indicates the similarity of two discretized probability distributions.
When the prediction of Xenovert is completely identical to a uniform distribution, it will receive a maximum value of 1.

\subsection{Result}

The result in Table.~\ref{tab5} revealed that Xenovert can achieve close to the maximum HI score consistently over the three types of distribution shifts with noticeable adaptivity.
This adaptivity is revealed by the HI score over time shown in Fig.~\ref{fig4}, in which the Xenovert can first adapt to the source distribution and reach equilibrium.
Once the shifting happens, the Xenovert starts adapting spontaneously to fit the target distribution, given enough input; this is indicated by the dropping and climbing of the HI score.
With the above-mentioned sufficient experiments and adequate results, we then generalized the approach to other applications, as described in the following sections.

Xenovert is further shown to be invariant to the degree of shifting. 
This is verified by the recorded identical HI scores of two experiments, conducted using (1) $\mathcal{N}(0, 5)$ and $\mathcal{N}(50, 5)$ as source distributions and (2) $\mathcal{N}(250, 5)$ as target distributions (Fig.~\ref{fig4}A).
Additionally, we analyzed the total levels of the Xenovert on the HI score (i.e., the number of quantized outputs).
The higher the total level, the lesser the data compression and information lost, as the number of quantized outputs increased.
The graph in Fig.~\ref{fig4}B shows that the HI score decreases as the total levels increase and data compression decreases.
Other than that, we investigated how the progression of the HI score was affected by different values of learning rate.
We observed that the higher the learning rate, the lower the stability but faster the adaptability (Fig.~\ref{fig4}C).
This is referred to as the stability-adaptability trade-off.
Note that an extreme learning rate (e.g., $0.01$) can cause large fluctuations, leading to a lower HI score.


\begin{table}[htbp]
\caption{HI scores for adapting to the shifting of Uniform, Normal, Multimodal Normal, and Chi-squared distributions, with different types of shifts. Result are mean ± SD.}
\begin{center}
\scalebox{0.8}{
\begin{tabular}{|c|c|c|c|}
\hline
\textbf{Distribution}&\textbf{Instant shift}&\textbf{Gradual shift}&\textbf{Recurring shift}\\
\hline
Uniform&$0.984 \pm 0.000$&$0.992 \pm 0.001$&$0.976 \pm 0.001$\\
Normal&$0.972 \pm 0.001$&$0.990 \pm 0.001$&$0.975 \pm 0.001$\\
Multimodal Normal&$0.982 \pm 0.001$&$0.984 \pm 0.001$&$0.989 \pm 0.001$\\
Chi-Squared&$0.983 \pm 0.000$&$0.981 \pm 0.001$&$0.984 \pm 0.001$\\
\hline
\end{tabular}
}
\label{tab5}
\end{center}
\end{table}



\section{Application on Covariate Shift Problem}

\subsection{Experiment setup}

Covariate shift refers to changes in the distribution of the input variables $x$, even when the mapping to the label $y$ remains unchanged in $X\rightarrow Y$ problems. 
In other words, it can be defined as the case where the conditional distribution $P_{tr}(y|x) = P_{tst}(y|x)$ remains unchanged but $P_{tr}(x) \neq P_{tst}(x)$.
Covariate shift mainly occurs under the influence of a non-stationary environment or sampling bias \cite{bib1D7, bib1D5, bib2D8, bib1D16}.
For instance, global climate change can be a driver of phenotypic changes in plants and crops.
Thus, the morphological shift can affect the measurement of their features over time \cite{bib2D9, bib2D10}.
This can potentially pose a problem for the classifier that failed to incorporate such an event.
Here, we investigate how our method can be applied to mitigate covariate shifts in real-world data.

\begin{figure}[htbp]
\centerline{\includegraphics[width = 7cm]{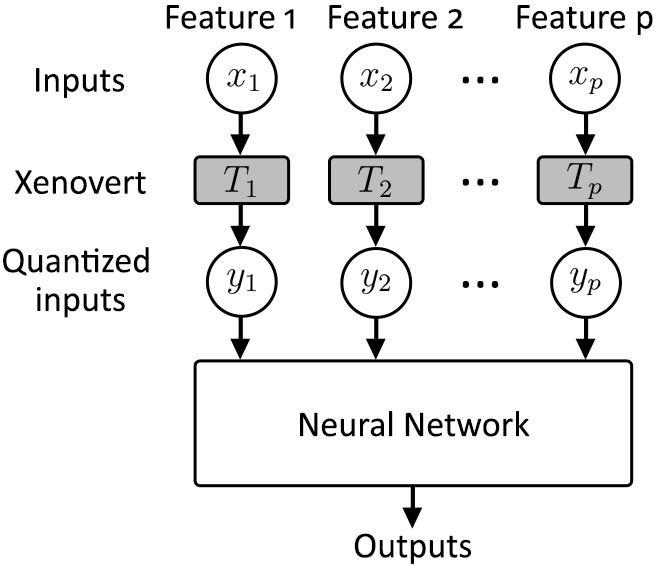}}
\caption{Diagram illustrating the integration of Xenoverts with a Neural Network. The dataset consists of $p$ features, and each feature is independently processed by one of the $p$ Xenoverts, resulting in a set of quantized inputs. These inputs are then fed into the neural network. As the Xenoverts adapt, the neural network is concurrently trained. Notably, after a covariate shift, the neural network's parameters can remain frozen, performing inference as usual. However, the Xenoverts continue to adapt to the evolving input distribution, ensuring the quantized inputs stay pertinent for the neural network's inference.}
\label{fig2}
\end{figure}

\begin{figure*}[h]
\centerline{\includegraphics[width = 18cm]{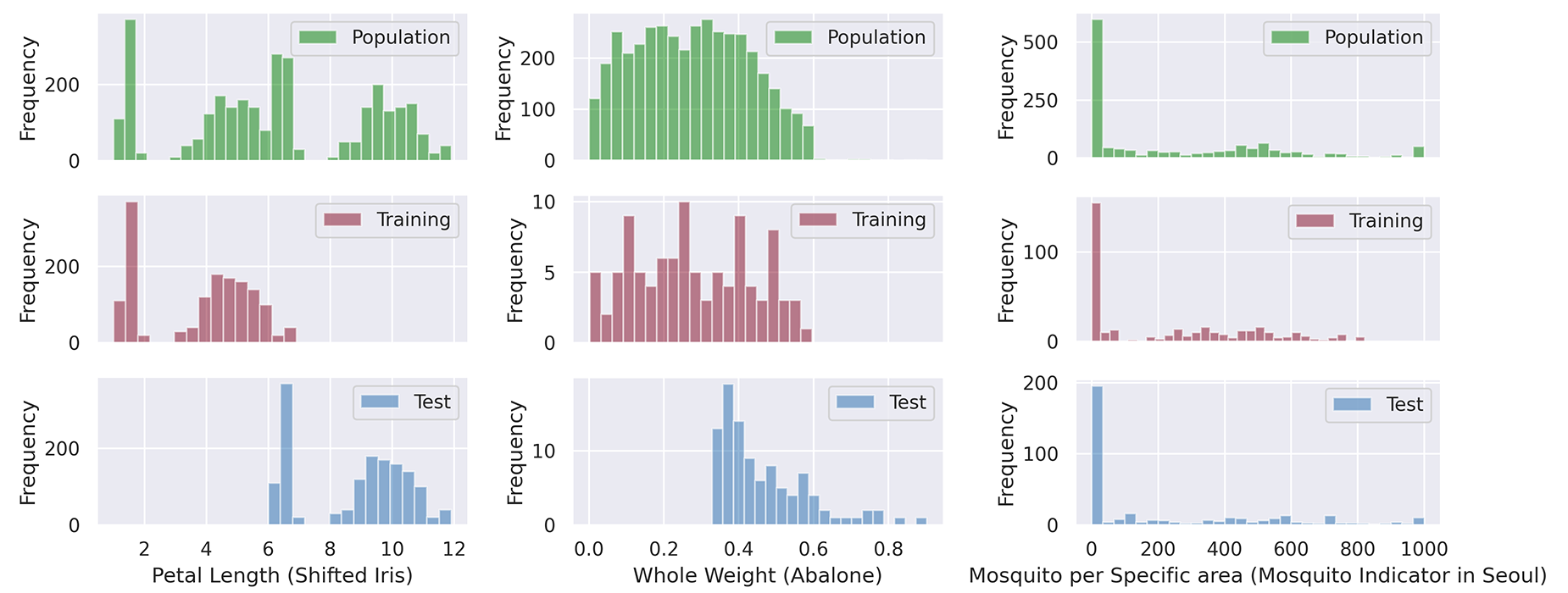}}
\caption{Feature distributions for three example datasets, illustrating the original population distribution and the corresponding shifted distributions for training and testing sets. The training set depicted is a subset, adjusted to match the size of the testing set for visualization purposes. These visualizations are referred from \cite{bib1D25}.}
\label{fig2_1}
\end{figure*}

\begin{figure}[htbp]
\centerline{\includegraphics[width = 9cm]{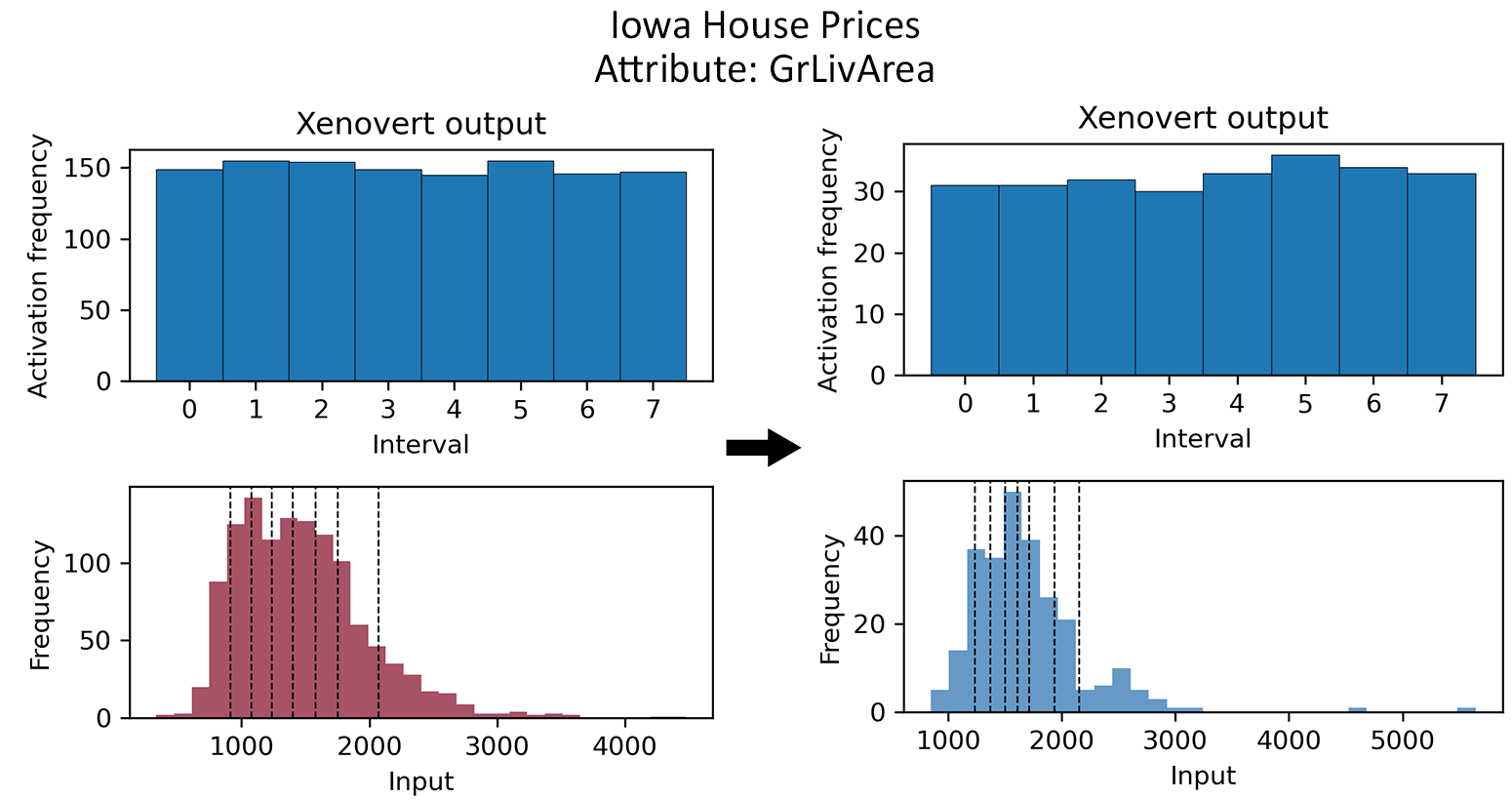}}
\caption{Adaptation of Xenovert to the input distribution before (left) and after (right) covariate shift. (Top) The activation frequencies of the intervals are kept approximately uniform online, achieved through Xenovert's reorganization of the values of its quasi-quantiles (dashed line).}
\label{fig2_2}
\end{figure}

This experiment compares neural networks trained under three conditions: with Xenovert, without Xenovert, and with DIW.
When integrated with Xenovert, the neural network employs the Xenovert(s) to transform raw inputs into quantized inputs based on the intervals they fall into (for the architecture of this integration, refer to Figure 2).
Every time an input is fed into the Xenovert, it will reconfigure its quasi-quantiles corresponding to the input.
A basic two-layer Multilayer Perceptron (MLP) was employed as the neural network model across all three conditions.
This simple MLP configuration primarily aimed to test the effects of Xenovert and DIW with minimal influence from the complexity of the neural network itself.

We conducted our experiments on five shifted datasets, comprising two for classification tasks and three for regression tasks.
The models' performances were evaluated using accuracy for the classification tasks and Mean Squared Error (MSE) for the regression tasks.
Our study includes real-world datasets, such as Diabetes \cite{bibdata2}, Abalone \cite{bibdata1}, Iowa House Prices \cite{bib4D2}, and the Mosquito Indicator in Korea. These datasets were chosen for their inherent natural shifts, including sampling bias and data rearranging before significant shifts occurred.
We also apply an artificial shift to the Iris dataset attributes to simulate a drastic shift, simulating the adaptive changes in plant attributes driven by global climate change, as mentioned earlier in this section.

The handling of each dataset, including the specific methods used for introducing shifts and preparing the data for analysis, is detailed in Appendix B-A.
Figure~\ref{fig2_1} provides examples of selected features from the datasets, illustrating their states before and after the shifts.
We also compared performance on unshifted datasets (original forms of Iris and Abalone) versus their shifted versions to evaluate any potential information loss caused by Xenovert.
For each dataset, we trained the standard MLP and the integrated models using the source distribution (pre-shift data) and evaluated their performance using the target distribution (post-shift data), assessing their ability to adapt to changes in data distribution.

\subsection{Result}


The results summarized in Table~\ref{tab1} show that MLP integrated with Xenovert achieved the best result in 4 out of 5 shifted datasets, especially one with drastic shifts.
This highlights Xenovert's ability to adapt to covariate shifts. 
In the Iris dataset, the standard MLP model exhibited a significant accuracy reduction from $1.00 \pm 0.00$ without shift to $0.36 \pm 0.67$ with shift.
In contrast, MLP with Xenovert effectively maintained its high accuracy, recording $0.99 \pm 0.01$ both with and without shift. 
For the Diabetes dataset, both the standard MLP and MLP with Xenovert showed a similar trend in terms of accuracy, respectively, whereas MLP with DIW had a lower accuracy.

In the shifted regression datasets, the standard MLP's performance on the Abalone dataset increased in MSE from $2.34 \pm 0.14$ in the unshifted condition to $7.038 \pm 2.635$ in the shifted condition.
MLP with Xenovert, however, demonstrated a lower MSE of $3.953 \pm 0.271$ under the shifted condition.
In the regression tasks across other tested datasets, MLP with Xenovert consistently demonstrated a lower MSE compared to the standard MLP.
This hints at its stable performance despite the influence of the covariate shift in regression tasks.
Note that MLP with DIW is not applicable for the regression tasks.

A separate analysis comparing performance on datasets with and without shifts reveals a minor degradation in performance for MLP with Xenovert.
This is likely due to information loss during the quantization process. 
Despite this, MLP with Xenovert demonstrated consistent resilience to shifts, maintaining stable performance metrics in both shifted and unshifted conditions. 
In contrast, the standard MLP model exhibited a significant decline in accuracy and an increase in MSE under shifted conditions, indicating a vulnerability to shifts in input, specifically those that are drastic.

\begin{table}[htbp]
\caption{Comparison of Neural Networks with Xenovert, without Xenovert, and with DIW on Covariate Shift Dataset.}
\begin{center}
\resizebox{\linewidth}{!}{
\begin{tabular}{l|c|c|c}
\hline
Dataset\textbackslash Method & MLP & MLP+Xenovert & MLP+DIW \\
\hline
\multicolumn{4}{l}{Classification (Accuracy $\uparrow$)} \\
\hline
Iris (no shift) & \bm{$1.00 \pm 0.00$} & $0.99 \pm 0.01$ & $0.98 \pm 0.01$ \\
Iris & $0.36 \pm 0.67$ & \bm{$0.99 \pm 0.01$} & $0.34 \pm 0.08$ \\
Diabetes & \bm{$0.56 \pm 0.07$} & \bm{$0.56 \pm 0.05$} & $0.52 \pm 0.09$ \\
\hline
\multicolumn{4}{l}{Regression (MSE $\downarrow$)} \\
\hline
Abalone (no shift) & \bm{$2.34 \pm 0.14$} & $2.41 \pm 0.11$ & - \\
Abalone & $7.038 \pm 2.635$ & \bm{$3.953 \pm 0.271$} & - \\
Iowa House Prices & $76199.49 \pm 3219.5$ & \bm{$61374.59 \pm 1169.3$} & - \\
Mosquito Indicator in Korea & $15.24 \pm 6.21$ & \bm{$13.99 \pm 1.05$} & - \\
\hline
\end{tabular}
}
\end{center}
\label{tab1}
\end{table}



\section{Computational Complexity Analysis}

We analyze the computational complexity of the two primary operations in the proposed Xenovert algorithm: the update function $f(x)$, which adapts quasi-quantiles to accommodate shifts in each input, and the convert function $C(x)$, which returns a quantized output.
Both functions exhibit a complexity of $O(N \cdot L)$, where $N$ is the number of inputs and $L$ is the depth of the tree.
The computational complexity of these functions is primarily due to their recursive traversal from the root to a leaf node in the tree by the depth $L$.


\section{Discussion}

In this study, we present an online method that adapts to distribution shifts, a critical challenge in dynamic data environments.
This is accomplished through the adaptive algorithm proposed by adjusting the value of quasi-quantiles to approximate the interval's average.

In the univariate distribution shift experiment, the Xenovert demonstrated its capability to adjust quasi-quantiles, guiding them towards the average within their intervals.
Consequently, it uniformly divided the input space, mapping any type of distribution into a uniform distribution.
Moreover, the results showed that Xenovert is robust against instant, gradual, and recurring distribution shifts.
Nonetheless, the Xenovert's output seldom produces a perfect uniform distribution as the most recent inputs account for the imperfection, since these types of adaptive algorithms are, arguably, more sensitive to recent temporal input.

Particularly, increasing the number of quasi-quantiles can reduce the information loss during compression when transforming intervals into quantized output.
However, this increase also heightens the sensitivity of each quasi-quantile, leading them to adjust their values more frequently, which consequently lowers the HI score.
In addition, the stability-adaptability trade-off hints that the learning rate should be tuned according to the shifting rate of the input distribution.
Heuristically, a problem with rapid shifting might require sacrificing precision for faster adaptability.
Conversely, the issue of having a task that requires more precision but is constrained by limited data can easily be mitigated by bootstrapping the data.

The experimental results show that integrating a neural network with the Xenovert enables adaptation to data with covariate shifts.
This can save the hurdle of periodically detecting shifts and retraining a classifier, especially for systems expected to be deployed in environments subject to change.
The key to achieving this is maintaining the mapping between the training input and the shifted testing input through the use of configurable quasi-quantiles. 
A notable limitation, however, is the potential information loss inherent in compressing raw input into a set of quantized units.
Therefore, the total level of the Xenovert needs to be scaled accordingly to balance information retention and quantization efficiency.
Our algorithm is not intended to supplant the current state-of-the-art methods for addressing covariate shift.
For example, methods like dynamic importance reweighting are more adept at handling shifts caused by noise or outliers, but less so with drastic shifts; 
Rather, we propose our algorithm as a simpler alternative for this specific issue while also offering a potential solution for other similar problems.

\section{Conclusion}
In conclusion, we may characterize Xenovert as a general-purpose adaptive algorithm that maintains an equal division of input distribution.
Due to its simplicity, we speculate the algorithm can be further applied to other applications.
Additionally, future work will focus on developing a more generalized version of Xenovert, capable of handling high-dimensional inputs, potentially an N-dimensional Xenovert.


%

\appendices
\section{Xenovert}

Algorithm~1 and Algorithm~2 displayed the pseudocode for the workflow of quasi-quantiles and Xenovert. 
The source code of the models has been made available to download in a DOI (doi.org/10.5281/zenodo.8127371).

\begin{algorithm}

\caption{Quasi-quantile}
\label{alg:alg1}
\begin{algorithmic}

\Class{QQ} \Comment{Quasi-quantile object}
    \State $leftChild :$ QQ 
    \State $rightChild :$ QQ
    \State\textit{$l = 0$} \Comment{Current level}
    \State\textit{$\alpha = 1e-05$} \Comment{Learning rate}
    \State\textit{$v = 0$} \Comment{Velocity}
    \State\textit{$\theta = 0.99$} \Comment{Velocity decay}
    \State\textit{$q = 0$} \Comment{Quasi-quantile's value}
\Statex
\Procedure{Grow}{$\alpha$, $q$, $l$}
    \If {$leftChild$ exist}
        \State $leftChild.\Call{Grow}{\alpha, q, l}$ \Comment{Call Grow recursively}
        \State $rightChild.\Call{Grow}{\alpha, q, l}$ 
    \Else
        \State\textit{Instantiate quasi-quantiles object as children}
        \State $leftChild :=$ QQ$(\alpha, q, l+1)$
        \State $rightChild :=$ QQ$(\alpha, q, l+1)$
    \EndIf
\EndProcedure
\Statex
\Procedure{Update}{$x$}    
    \State $v' = \theta \times v + |q-x|$
    \State $s = 1$
    \If {$ q-x > 0 $}
        \State $s = -1$
    \EndIf

    \State $q' = q + \alpha \times v' \times s$

    \If {$leftChild$ exist}
        \If {$ x < q$}
            \State $leftChild$.\Call{Update}{$x$}
        \Else
            \State $rightChild$.\Call{Update}{$x$}
        \EndIf
    \EndIf

\EndProcedure
\Statex
\State\textit{Output a quantized output according to the selected interval}
\Procedure{Convert}{$x$, \textit{offset}}    
    \If {$leftChild$ exist}
        \If {$ x < q$}
            \State $leftChild$.\Call{Convert}{$x$, \textit{offset}}
        \Else
            \State $rightChild$.\Call{Convert}{$x$, $2^l+$\textit{offset}}
        \EndIf
    \Else
        \If {$ x < q$}
            \State \textbf{return} \textit{offset}
        \Else
            \State \textbf{return} $2^l+$\textit{offset}
        \EndIf
        
    \EndIf

\EndProcedure
\EndClass
\end{algorithmic}
\end{algorithm}

\begin{algorithm}
\caption{Xenovert}
\label{alg:alg2}
\begin{algorithmic}

\Class{Xenovert}  \Comment{Xenovert object}
    \State\textit{$\alpha = 1e-05$} \Comment{Learning rate}
    \State\textit{$q = 0$} \Comment{Root quasi-quantile's value}
    \State $rootQuasiQuatile := $ QQ($\alpha$, $q$) \Comment{Root quasi-quantile}
    \State\textit{$X$} $\leftarrow$ Input distribution
\Statex
\State\textit{Grow the binary tree $L$ times}
\State $l = 0$
\For {$L$}
    \State $rootQuasiQuatile.\Call{Grow}{\alpha, q, l}$ 
\EndFor
\Statex
\Procedure{Update}{$X$}
    \For {$x \in X$}
        \State $rootQuasiQuatile.\Call{Update}{x}$
    \EndFor
\EndProcedure
\Statex
\State\textit{Output a quantized output based on input $x$ }
\Procedure{Output}{$x$}
    \State \textbf{return} $rootQuasiQuatile.\Call{Convert}{x}$
\EndProcedure

\EndClass
\end{algorithmic}
\end{algorithm}



\section{Experiment details}

\subsection{Covariate shift Problem}
\label{sec:Covariat- shift-Problem}
In our experiment, we prepared five shifted datasets, including synthetic datasets and real-world datasets, to explore the impact of covariate shift on machine learning model performance. Here we denote the training set as $X_{tr}$ and shifted testing set as $X_{te}$, with each dataset split based on the predefined criteria specified below. Note that we added Gaussian noise to the input features if the attribute is an integer to prevent the true divide issue. Xenovert can be iteratively updated with online data, but this can introduce some bias to the most recent data if they are temporally correlated over many samples. To mitigate this, we can implement shuffles on the data distribution that is fed into Xenovert.

We compare two neural networks that learn with Xenovert and without Xenovert.
When working with Xenovert, each i.i.d feature in $X_{tr}$ will be fed into a corresponding Xenovert for the adaptation process.
For instance, in the case of the Iris dataset, 4 features are paired with 4 Xenoverts.
Following that, each Xenovert performed adaptation and produced a quantized output $O \in \{0, M\} \in \mathbb{Z}$.
The quantized outputs are used as raw inputs to train a Multi-layer perceptron network (MLP) for the downstream operation.
The architecture of the network is composed of two hidden layers; each layer contains 200 neurons and a ReLU activation function.
The input layer takes four inputs, and the output layer outputs the cross entropy of three classes.
For the MLP training, we set the batch size to 200, the epoch to 2000, and a learning rate of 0.01.
After the network is trained, we feed $X_{te}$ to Xenoverts for adaptation and to produce quantized outputs.
The network takes those quantized outputs and performs classification without retraining.
On the other hand, the neural network without Xenovert, which adopted the same architecture, is trained with a training set and performs classification on shifted testing sets without re-training.
We then compare their accuracy to determine the effectiveness of Xenovert in handling covariate shift.

\subsubsection{Iris \cite{bib4D1}}
The Iris dataset preprocessing introduced an artificial shift by increasing petal length and width in $X_{te}$ by 5 units. The original dataset served as the $X_{tr}$, creating a distinct morphological shift of the species.

\subsubsection{Diabetes \cite{bibdata2}}
We order the dataset by age and split it by patients younger than age 24 as $X_{tr}$ and patients older and equal to age 24 as the $X_{te}$. We select \textit{Glucose} and \textit{BMI} as features, determined by Pearson’s correlation to predict whether a patient has diabetes.

\subsubsection{Abalone \cite{bibdata1}}
We apply sampling bias by segmenting based on the feature \textit{Whole weight} of the abalones, following the experiment setting from \cite{bib1D25}.

\subsubsection{Iowa House Prices \cite{bib4D2}}
We split the dataset by the feature \textit{YearBuilt}, which indicates the year the house was built. Data of houses that were built before or during the year 2000 will be included in $X_{tr}$. Data of houses that were built after the year 2000 will be included in $X_{tr}$. We selected \textit{GrLivArea} and \textit{OverallQual} as features, determined by Pearson’s correlation to the label, house prices.

\subsubsection{Mosquito Indicator in Korea}
We introduce sampling bias through temporal partitioning of the data, designating records up until 2018 for $X_{tr}$ and those from 2019 forward for $X_{te}$.


\section{Statistical analysis}

The reported statistical test was analyzed using Python (SciPy). The data shown in the covariate shift study were obtained from at least 30 independent experiments. The results gathered in the univariate experiments were obtained from at least 30 independent experiments. Values in different experimental groups are expressed as the mean ± standard deviation. $p < 0.05$ was considered statistically significant.

\section*{Acknowledgment}

This work was supported by JST, ACT-I Grant Number JP-50243 and JSPS KAKENHI Grant Number JP20241216. Tham Yik Foong and Heng Zhang are supported by JST SPRING, Grant Number JPMJSP2136.

\ifCLASSOPTIONcaptionsoff
  \newpage
\fi



%
\bibliographystyle{IEEEtran} 
\bibliography{main,heng}

%

\newpage

\begin{IEEEbiography}[{\includegraphics[width=1in,height=1.25in,clip,keepaspectratio]{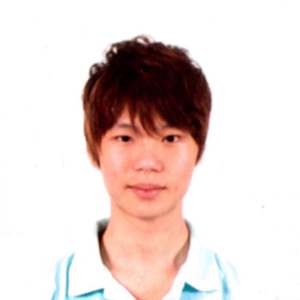}}]{Tham Yik Foong}
received a BSc (Hons) in Software Engineering with Multimedia from Limkokwing University in 2017. He then received an MSc in Machine Learning in Science at the University of Nottingham in 2020. Currently, he is working towards his Ph.D. degree at Kyushu University, Japan, since 2021, supported by the prestigious SPRING Scholarship from one of Japan’s largest funding agencies. his research interests lie in adaptive algorithms, reinforcement learning, and reservoir computing.
\end{IEEEbiography}

\begin{IEEEbiography}[{\includegraphics[width=1in,height=1.25in,clip,keepaspectratio]{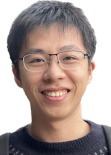}}]{Heng Zhang}
received the B.Eng. degree from the Department of Electronic Information Engineering, Shenzhen University, China, 2019, and the M.Sc. degree from the Department of Electrical and Electronic Engineering, The University of Manchester, 2020. He is currently pursuing the Ph.D. degree in Kyushu University, Japan, since 2021. His current research interests include bio-inspired machine learning, artificial intelligence, and robust and adaptive learning algorithms. He has been receiving the SPRING Scholarship to study at Kyushu University since 2021, which is supported by one of the biggest Japanese funding agencies.
\end{IEEEbiography}


\begin{IEEEbiography}[{\includegraphics[width=1in,height=1.25in,clip,keepaspectratio]{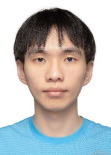}}]{Mao Po Yuan}
received a BSc in Mechanical Engineering from National Chung Hsing University in 2020. Currently, he is working towards his master’s degree at Kyushu University, Japan, since 2022. His research interests lie in generative model, computer vision, and robustness.
\end{IEEEbiography}

\begin{IEEEbiography}[{\includegraphics[width=1in,height=1.25in,clip,keepaspectratio]{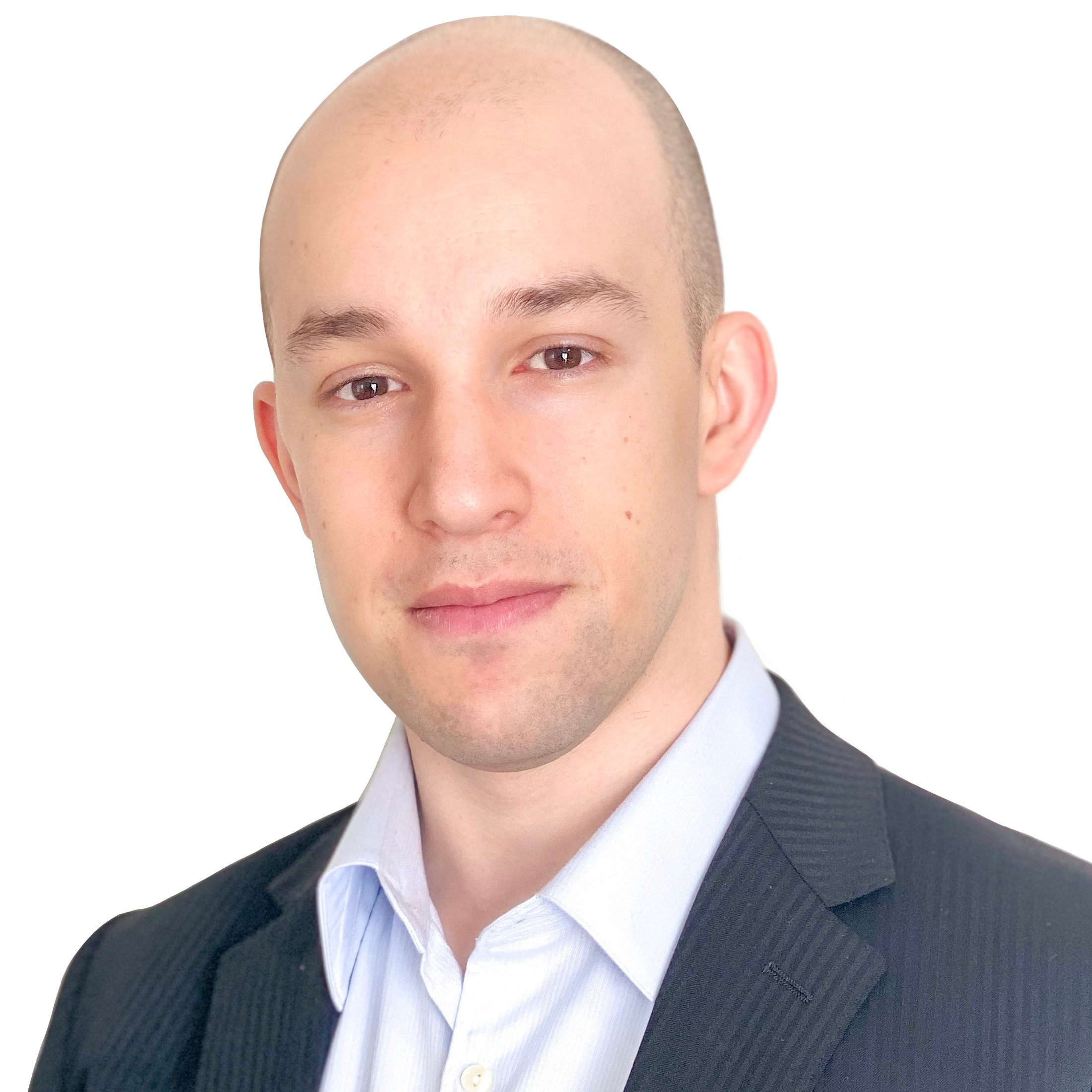}}]{Danilo Vasconcellos Vargas}
is an associate professor at Kyushu University, a visiting researcher at the University of Tokyo, and CEO/Founder of MiraiX. His research interests include AI, evolutionary computation, and complex adaptive systems. He has published numerous works in prestigious journals and received several awards, including the IEEE Transactions on Evolutionary Computation Outstanding 2022 Paper award. He has presented tutorials at various conferences, co-organized and advised workshops, and leads the Laboratory of Intelligent Systems focused on building robust and adaptive AI funded by JST and JSPS.
\end{IEEEbiography}




\end{document}